\documentclass{article}
\usepackage[nonatbib, final]{neurips_2019}

\usepackage[utf8]{inputenc} %
\usepackage[T1]{fontenc}    %
\usepackage{hyperref}       %
\usepackage{url}            %
\usepackage{booktabs}       %
\usepackage{amsfonts}       %
\usepackage{nicefrac}       %
\usepackage{microtype}      %

\usepackage{microtype}
\usepackage{graphicx}
\usepackage{subfigure}
\usepackage{booktabs} %
\usepackage{amsmath, amsthm, amssymb, amsfonts}

\usepackage{algorithm}
\usepackage{algpseudocode}

\usepackage{hyperref}

\usepackage{todonotes}

\title{Model Weight Theft With Just Noise Inputs: \\The Curious Case of the Petulant Attacker}

\author{%
  Nicholas Roberts, Vinay Uday Prabhu, Matthew McAteer  \\
  UnifyID AI Labs\\
  \texttt{ncrobert@cs.cmu.edu, \{vinay, matthew\}@unify.id} \\
}

\begin{document}

\maketitle

\begin{abstract}
This paper explores the scenarios under which an attacker can claim that `Noise and access to the softmax layer of the model is all you need' to steal the weights of a convolutional neural network whose architecture is already known. We were able to achieve $96\%$ test accuracy using the stolen MNIST model and $82\%$ accuracy using the stolen KMNIST model learned using only i.i.d. Bernoulli noise inputs. We posit that this theft-susceptibility of the weights is indicative of the complexity of the dataset and propose a new metric that captures the same. The goal of this dissemination is to not just showcase how far knowing the architecture can take you in terms of model stealing, but to also draw attention to this rather idiosyncratic weight learnability aspects of CNNs spurred by i.i.d. noise input. We also disseminate some initial results obtained with using the Ising probability distribution in lieu of the i.i.d. Bernoulli distribution. 
\end{abstract}

\section{Introduction}
In this paper, we consider the fate of an attacker who is adamant about only using noise as input to a convolutional neural network (CNN) whose architecture is known and whose weights are the target of theft. We assume that the attacker has earned access to the softmax layer and is not restricted in terms of the number of inputs to be used to carry out the attack.
\\At the outset, we'd like to emphasize that our goal in disseminating these results is not to convince the reader on the real-world validity of the attacker-scenario described above or to showcase a novel attack. This paper contains our initial explorations after a chance discovery that we could \textit{functionally replicate} the weights of an MNIST-trained CNN model by just using noise as input into the framework described below.
\\Through a set of empirical experiments, which we are duly open sourcing to aid reproducibility, we seek to draw the attention of the community on the following two issues:

\begin{enumerate}
\item This risk of model weight theft clearly entails an interplay between the dataset as well as the architecture. Given a fixed architecture, can we use the level of susceptibility as a novel metric of complexity of the dataset?
\item Given the wide variations in success attained by varying the noise distribution, how do we formally characterize the relationship between the input noise distribution being used by the attacker and the true distribution of the data, while considering a specific CNN architecture? What aspects of the true data distribution are actually important for model extraction? 
\end{enumerate}
The rest of the paper is structured as follows:
\\In Section 2, we provide a brief literature survey of the related work. In Section 3, we describe the methodology used to carry out the attack. In Section 4, we cover the main results obtained and conclude the paper in Section 5.

\section{Related work}
The art form of \textit{stealing} machine learning models has received a lot of attention in the recent years. In \cite{stealing_tramer}, the authors specifically targeted real-world ML-as-a-service \cite{mlaas} platforms such as BigML and Amazon Machine Learning and demonstrated effective attacks that resulted in extraction of machine learning models with \textit{near-perfect fidelity} for several popular model classes. In \cite{copycat_cnn}, the authors trained what they termed as a \textit{copycat network} using \textit{Non-Problem Domain} images and stolen labels to achieve impressive results in the three problems of facial
expression, object, and crosswalk classification. This was followed by work on \textit{Knockoff Nets} \cite{knockoff_net}, where the authors demonstrated that by merely querying with random images sourced from an entirely different distribution than that of the black box target training data, one could not just train a well-performing knockoff but it was possible to achieve high accuracy even when the knockoff  was constructed using a completely different architecture.
\\This work differs from the above works in that the attacker is adamant on only using \textit{noise} images as querying inputs. Intriguingly enough, the state-of-the-art CNNs are not robust enough to provide a flat (uniform) softmax output (with weight $1/\textit{number-of-classes}$) when we input non-input-domain noise at the input layer. This has been studied under two contexts. The first context was within the framework of \textit{fooling images}. In \cite{nguyen2015deep}, the authors showcased how to generate synthetic images that were noise-like and completely unrecognizable to the human-eye but ones that state-of-the-art CNNs classified as one of the training classes with $99.99\%$ confidence. The second text was with regards to what the authors in \cite{goodfellow2014explaining} stated to be \textit{rubbish-class examples} . Here, they showcased that the high levels of confident mis-predictions exuded by state-of-the-art trained on MNIST and CIFAR-10 datasets in response to isotropic Gaussian noise inputs.
\\ In this work, we focus on using Bernoulli noise-samples as inputs and using the softmax responses of the target model to siphon away the weights.
\section{Methodology}

\subsection{Threat model}
We propose a framework for model extraction without possession of samples from the true dataset which the model has been trained on or the purpose of the model other than the dimensionality of the input tensors as well as the ability to access the resulting class distribution from what is assumed to be a softmax activation given an input. We make the additional assumption that the architecture of the model to be extracted is known by the adversary. In our experiments, we assume that the input tensor is of dimension $28$ by $28$ and each pixel has values on the interval $[0, 1]$.

\subsection{Victim model}
The black box model which we attempt to extract, $F(\cdot)$, whose architecture is described in Table~\ref{table:model-bb-arch}, is trained to convergence on a standard dataset for $12$ epochs using the Adadelta optimizer with an initial learning rate of 1.0 and a minibatch size of $128$ \cite{Mnistcnn1:online}. From this point onward, this model is assumed to be a black box in which we have no access to the parameters of each layer. 

\begin{table}[t]
\caption{Victim architecture as found in the MNIST example in the documentation for the Keras deep learning library.}
\label{table:model-bb-arch}
\vskip 0.15in
\begin{center}
\begin{small}
\begin{sc}
\begin{tabular}{lccr}
\toprule
Layer type    & Dimensions & Additional \\&&information   \\
\midrule
Convolutional & $32$, $3 \times 3$ & ReLU                   \\
Convolutional & $64$, $3 \times 3$ & ReLU                   \\
Max Pooling   &       $2 \times 2$ & -                      \\
Dropout       &                    & $\text{rate} = 0.25$   \\
Dense         & $128$              & ReLU                   \\
Dropout       & -                  & $\text{rate} = 0.5$    \\
Dense         & $10$               & Softmax                \\
\bottomrule
\end{tabular}
\end{sc}
\end{small}
\end{center}
\vskip -0.1in
\end{table}

\subsection{Random stimulus response for model extraction}
We procedurally generate a dataset of `stimuli' comprised of $600000$ $28$ by $28$ binary tensors where each pixel is sampled from a Bernoulli distribution with a success probability parameter $p$. In other words, let each image $x_{\textit{rand}}^{i} \in X_{\textit{rand}} \subseteq \{0, 1\}^{28 \times 28}$ where $x_{\textit{rand}, j, k}^{i} \sim \operatorname{Bern}(p)$ for $i \in \{1, ..., 600000\}$. We sample these tensors with probability parameters $p \in \{0.01, 0.11, ... 0.91\}$, where each $p$ is used to generate $10\%$ of the data. We obtain predictions from the black box model for each randomly sampled example, $y_{\textit{rand}}^{i} = F(x_{\textit{rand}}^{i})$, which we refer to as `responses.' 

\subsection{Extraction}
\begin{figure}[ht]
  \centering
  \includegraphics[height=0.4\textwidth]{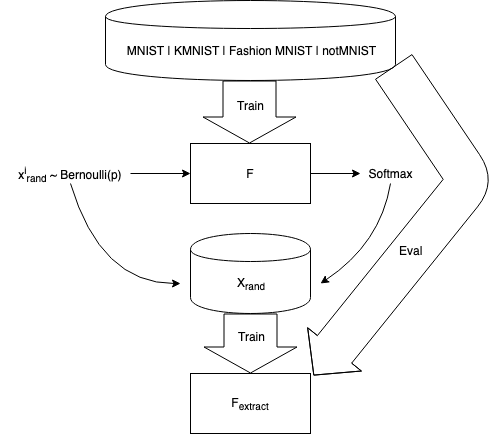} 
  \caption{Overview of the model extraction algorithm.}
  \vspace{-0.15in}
   \label{fig:alg-summary}
\end{figure}

We train a new model, $F_{\textit{extract}}(\cdot)$, on the stimulus response pairs, $\{(x_{\textit{rand}}^{i}, y_{\textit{rand}}^{i})\}_{i=1}^{600000}$ pairs with no regularization and evaluate on the dataset originally used to train $F(\cdot)$. The architecture for this model is the same as $F(\cdot)$, except we remove the dropout layers to encourage overfitting. We train for 50 epochs using the Adadelta optimizer with an initial learning rate of 1.0 and a minibatch size of $128$. Additionally, we acknowledge a significant class imbalance in the highest probability classes in the softmax vectors $y_{\textit{rand}}$, so we remedy this by computing class weights according to the $\operatorname{argmax}$ of each softmax vector, and applying this re-weighting during the training of $F_{\textit{extract}}(\cdot)$. We show the full extraction algorithm in Algorithm~\ref{alg:stimulus-resp} and summarize it in Figure~\ref{fig:alg-summary}. 

We evaluate our proposed framework on four datasets from the MNIST family of datasets with identical dimensions: MNIST, KMNIST, Fashion MNIST, and notMNIST \cite{lecun-mnisthandwrittendigit-2010, clanuwat2018deep, xiao2017/online, notMNIST66:online}. 

\subsection{Experiments with noise distributions}

We evaluated the effect of sampling random data $x_{\textit{rand}}^{i}$ from different distributions on the performance of $F_{\textit{extract}}(\cdot)$ on the MNIST validation set. We used the same training procedure as found in the previously described experiments with two exceptions: we sample only $60000$ procedurally generated examples and we train $F_{\textit{extract}}(\cdot)$ for only 10 epochs. We evaluated the use of the uniform distribution on the bounded interval $[0, 1]$, the standard normal distribution, the standard Gumbel distribution, the Bernoulli distribution with success parameter $p=0.5$, and samples from an Ising model simulation with inverse temperature parameter $\beta \in [0.0, 0.1, ..., 0.9]$ and resulting values scaled to $\{0, 1\}$. 

\subsection{The Ising prior  as a model of spatial correlation}
The Ising prior is defined by the density \cite{taroni2015statistical}:
\[p\left( {\mathbf{x}} \right) = \frac{{\exp \left[ { - \beta \sum\limits_{ij \in E} {\left( {{x_i}{x_j}} \right)} } \right]}}{{\sum\limits_{\mathbf{x}} {\exp \left[ { - \beta \sum\limits_{ij \in E} {\left( {{x_i}{x_j}} \right)} } \right]} }};{x_i} \in \left\{ {-1,1} \right\}\]
Examples of images sampled from the Ising model can be found in Figure~\ref{fig:ising_examples}. 

For this experiment, we evaluated the role of the inverse temperature parameter, $\beta$,  of the Ising sampler in training $F_{\textit{extract}}(\cdot)$. We first partition the stimulus response pairs, $(X_{\textit{Ising}}, Y_{\textit{Ising}})$ into $10$ subsets with $7000$ examples each corresponding to the different $\beta$ parameters used to generate the samples, where $(X_{\textit{Ising}}, Y_{\textit{Ising}}) = \bigcup_{\beta \in \{ 0.0, 0.1, ..., 0.9 \}} \{(X_{\textit{Ising}, \beta}, Y_{\textit{Ising}, \beta})\}$. We train $F_{\textit{extract}}(\cdot)$ for $10$ epochs for each $\beta$ and validate on the original dataset. We performed this experiment for MNIST, KMNIST, Fashion MNIST, and notMNIST and report the variation in performance over different values of $\beta$.  

\begin{algorithm}[tb]
   \caption{Stimulus response model extraction.}
   \label{alg:stimulus-resp}
\begin{algorithmic}
   \State {\bfseries Input:} data $X_{\textit{train}}$, $Y_{\textit{train}}$, $X_{\textit{val}}$, $Y_{\textit{val}}$
   \State Initialize $F(\cdot)$.
   \State Initialize $\textit{numRandomExamples} = 600000$. 
   \State Initialize $\textit{dim} = 28$. 
   \State Fit $F(X_{\textit{train}})$, $Y_{\textit{train}}$.
   \State Evaluate $F(X_{\textit{val}})$, $Y_{\textit{val}}$.
   
   \For{$p$ in $\{0.01, 0.11, ..., 0.91\}$}
   \For{$q$ in $\{0, 1, ..., \textit{numRandomExamples}/10\}$}
   \For{$j$ in $\{0, 1, ..., \textit{dim-}1\}$}
   \For{$k$ in $\{0, 1, ..., \textit{dim-}1\}$}
   \State $x_{\textit{sample}, j, k} \sim \operatorname{Bern}(p)$
   
   \EndFor
   \EndFor
   \State $X_{\textit{rand}} = X_{\textit{rand}} \cup \{x_{\textit{sample}}\}$
   \EndFor
   \EndFor
   
   \State Initialize $F_{\textit{extract}}(\cdot)$.
   \For{$i \in \{1, ..., |X_{\textit{rand}}|\}$} 
   \State $y_{\textit{rand}}^i = F_{\textit{extract}}(x_{\textit{rand}}^i)$
   \EndFor
   \State Compute class weights $CW_{Y\textit{rand}}$ given $Y_{\textit{rand}}$
   \State Fit $F_{\textit{extract}}(X_{\textit{rand}})$, $Y_{\textit{rand}}$ with $CW_{Y\textit{rand}}$.
   \State Evaluate $F_{\textit{extract}}(X_{\textit{val}})$, $Y_{\textit{val}}$.

\end{algorithmic}
\end{algorithm}

\section{Results}

\subsection{MNIST}

We evaluate the efficacy of our framework by training $F(\cdot)$ on MNIST and going on to evaluate the performance of $F_{\textit{extract}}(\cdot)$ on MNIST after extraction. We found that $F(\cdot)$ achieved a validation accuracy of $99.03\%$ and $F_{\textit{extract}}(\cdot)$ achieved a validation accuracy of $95.93\%$. The distribution of the $\operatorname{argmax}$ of $Y_{\textit{rand}}$ can be found in Figure~\ref{fig:randclass-dist}. The most underrepresented class according to the $\operatorname{argmax}$ of $Y_{\textit{rand}}$ was class 6 represented by $198$ out of $600000$ random examples.

\subsection{KMNIST}

Our experiments with KMNIST resulted in $F(\cdot)$ achieving a validation accuracy of $94.79\%$ and $F_{\textit{extract}}(\cdot)$ achieving a validation accuracy of $81.18\%$. Class 8 was found to be the class with the fewest representatives according to the $\operatorname{argmax}$ of $Y_{\textit{rand}}$, which had $272$ representative examples out of $600000$.

\subsection{Fashion MNIST}

On the Fashion MNIST dataset, we found that $F(\cdot)$ achieved a validation accuracy of $92.16\%$, while $F_{\textit{extract}}(\cdot)$ achieved a validation accuracy of $75.31\%$. For Fashion MNIST, the most underrepresented class according to the $\operatorname{argmax}$ of $Y_{\textit{rand}}$ was class 7 (sneaker) with only 12 out of $600000$ random examples. Notably, the most common mispredictions according to Figure~\ref{fig:randclass-conf} were incorrectly predicting class 5 (sandal)  when the ground truth is class 7 (sneaker) and predicting class 5 (sandal) when the ground truth is class 9 (ankle boot). $F_{\textit{extract}}(\cdot)$ seems to predict the majority of examples from shoe-like classes to be of class 5 (sandal). 

\subsection{notMNIST}

We found that the notMNIST dataset had a more uniform class distribution according to the $\operatorname{argmax}$ of $Y_{\textit{rand}}$ than the other datasets that we evaluated. The class with the fewest representatives in this sense was class 9 (the letter j) with $3950$ out of $600000$ examples. Despite this potential advantage, the extracted model $F_{\textit{extract}}(\cdot)$ failed to generalize to the notMNIST validation set, achieving an accuracy of $10.47\%$, and as can be seen in Figure~\ref{fig:randclass-conf}, $F_{\textit{extract}}(\cdot)$ predicts class 5 (the letter e) in the vast majority of cases. In contrast, $F(\cdot)$ achieved a validation accuracy of $88.62\%$. 

\subsection{The performance of different noise distributions}

In evaluating the effect of sampling from different distributions to construct $X_{\textit{rand}}$, we found that among the uniform, standard normal, standard Gumbel, Bernoulli distributions, and the Ising model, samples from the Ising model attained the highest accuracy at $98.02\%$ when evaluating $F_{\textit{extract}}(\cdot)$ on the MNIST validation set. The results for each of the other distributions can be found in Table~\ref{table:noisedists}. We postulate that this is due to the modelling of spatial correlations, which is a property which is lacking when sampling from the uniform, standard normal, standard Gumbel, and Bernoulli distributions, as the pixels are assumed to be i.i.d.

\begin{table}[t]
\caption{Performance using different noise distributions.}
\label{table:noisedists}
\vskip 0.15in
\begin{center}
\begin{small}
\begin{sc}
\begin{tabular}{lcr}
\toprule
Distribution    & $F_{\textit{extract}}(\cdot)$ \\&validation accuracy   \\
\midrule
Uniform ($a=0, b=1$)                & $11.72\%$ \\
Standard Normal & \\
($\mu=0, \sigma=1$)                  & $68.79\%$ \\
Standard Gumbel & \\
($\mu=0, \beta=1$)                  & $70.03\%$ \\
Bernoulli ($p=0.5$)                 & $76.58\%$ \\
Ising ($\beta \in \{0.0, 0.1, ..., 0.9\}$) & $98.02\%$ \\
\bottomrule
\end{tabular}
\end{sc}
\end{small}
\end{center}
\vskip -0.1in
\end{table}

\subsection{Extraction hardness resulting from data}

We propose a measure of model extraction hardness resulting from the dataset which the original model is trained on as the ratio of the post-extraction validation accuracy (using $F_{\textit{extract}}(\cdot)$) and the pre-extraction validation accuracy (using $F(\cdot)$) under our framework. We show that the resulting ratios are align with the mainstream intuition regarding the general relative learnability of MNIST, KMNIST, Fashion MNIST, and notMNIST. For MNIST, we found this ratio to be $0.9687$, the ratio for KMNIST was $0.8564$, for Fashion MNIST we found it to be $0.8171$, and notMNIST achieved a ratio of $0.1181$. 

\subsection{The role of modelling spatial correlation}

We found that the loss and accuracy demonstrate an Occam's hill effect when the value of $\beta$ is varied, which, as Figure~\ref{fig:ising_acc} demonstrates, is particularly clear in the cases of MNIST and KMNIST. In Figure~\ref{fig:ising_loss}, we see that across datasets, the losses tend to be minimized around $\beta = 0.3$, however the behavior of larger values of $\beta$ varies from dataset to dataset. We postulate that this is indicative of the different distributions of the amount of spatial correlation across each dataset. We also found that accuracy is maximized at $\beta = 0.3$ for MNIST, KMNIST, and Fashion MNIST. We found that the optimal setting for $\beta$ for notMNIST was $\beta = 0.8$, where the behavior here varies as $\beta$ increases from the optimal value. 

\begin{table}[t]
\caption{Performance on original dataset before and after extraction (measured on the validation set).}
\label{table:perf}
\vskip 0.15in
\begin{center}
\begin{small}
\begin{sc}
\begin{tabular}{lccr}
\toprule
Dataset     &    Pre-extraction & Post-extraction \\& accuracy& accuracy   \\
\midrule
MNIST           & $99.03\%$ & $95.93\%$ \\
KMNIST          & $94.79\%$ & $81.18\%$ \\
Fashion \\MNIST & $92.16\%$ & $75.31\%$ \\
notMNIST        & $88.62\%$ & $10.47\%$ \\
\bottomrule
\end{tabular}
\end{sc}
\end{small}
\end{center}
\vskip -0.1in
\end{table}

\section{Conclusion and future work}

In this paper, we demonstrated a framework for extracting model parameters by training a new model on random impulse response pairs gleaned from the softmax output of the victim neural network. We went on to demonstrate the variation in model extractability based on the dataset which the original model was trained on. Finally, we proposed our framework as a method for which relative dataset complexity can be measured. 

\subsection{Future work}
This is a work in progress and we are currently working along the following three directions:
In our experiments, pixels are notably i.i.d., whereas in real world settings, image data is comprised of pixels which are spatially correlated. In this vein, we intend to establish the relationship between the temperature of an Ising prior and the accuracy obtained by the stolen model.
We will experiment with different architectures, specifically exploring the architecture unknown scenario where the attacker has a fixed plug-and-play swiss-army-knife architecture whose weights are learned by the noise and true-model softmax outputs. 
Additionally, we will explore methods for constructing $X_{\textit{rand}}$ which gives more uniform distributions over $\operatorname{argmax}(Y_{\textit{rand}}$) and evaluate the associated effect on the performance of $F_{\text{extract}}(\cdot)$.

\bibliography{ref}
\bibliographystyle{plain}

\appendix
\section{Additional figures}

\begin{figure}[ht]
  \centering
  \includegraphics[height=0.22\textwidth]{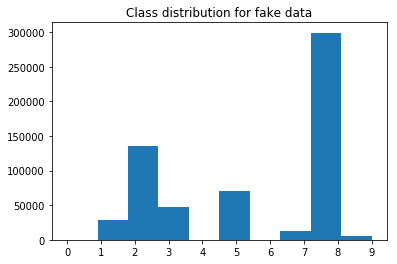}  \\
  \includegraphics[height=0.22\textwidth]{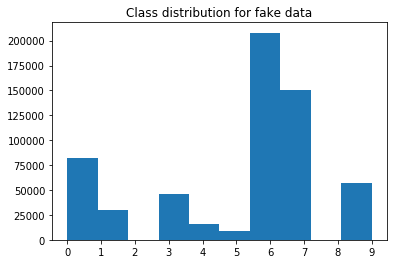} \\ 
  \includegraphics[height=0.22\textwidth]{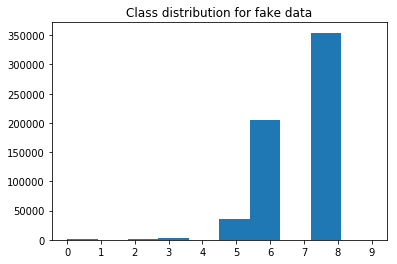} \\
  \includegraphics[height=0.22\textwidth]{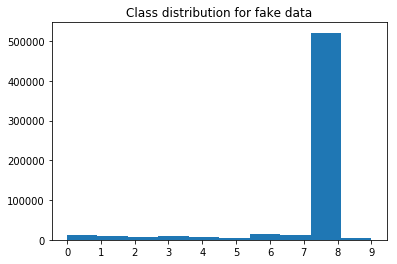}
  \caption{Distribution of classes given $X_{\textit{rand}}$. From top to bottom: MNIST, KMNIST, Fashion MNIST, notMNIST.}
  \vspace{-0.15in}
   \label{fig:randclass-dist}
\end{figure}

\begin{figure}[ht]
  \centering
  \includegraphics[height=0.22\textwidth]{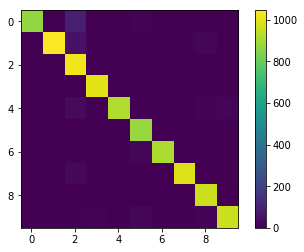}  \\
  \includegraphics[height=0.22\textwidth]{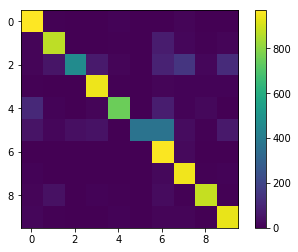} \\ 
  \includegraphics[height=0.22\textwidth]{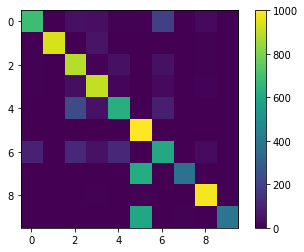} \\
  \includegraphics[height=0.22\textwidth]{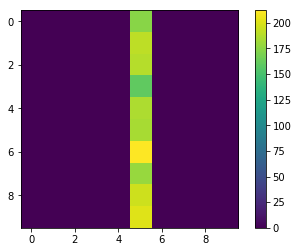}
  \caption{Confusion matrices of $F_{\textit{extract}}(\cdot)$ on $X_{\textit{val}}$. From top to bottom: MNIST, KMNIST, Fashion MNIST, notMNIST.}
  \vspace{-0.15in}
   \label{fig:randclass-conf}
\end{figure}

\begin{figure}[ht]
  \centering
   \begin{tabular}{cccc} 
  $\beta = 0.0$&
  \includegraphics[height=0.1\textwidth]{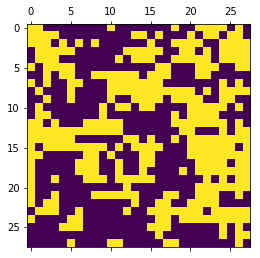} &
  \includegraphics[height=0.1\textwidth]{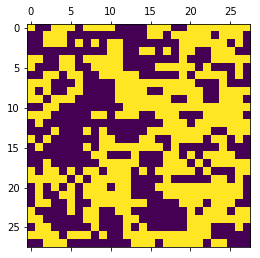} &
  \includegraphics[height=0.1\textwidth]{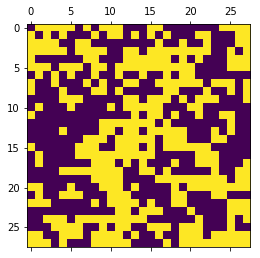} \\
  
  $\beta = 0.1$&
  \includegraphics[height=0.1\textwidth]{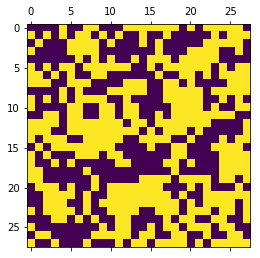} &
  \includegraphics[height=0.1\textwidth]{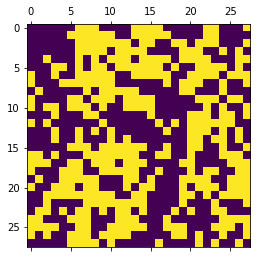} &
  \includegraphics[height=0.1\textwidth]{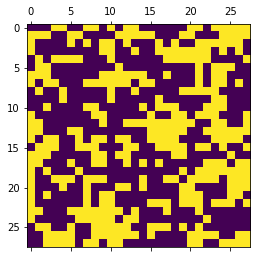} \\
  
  $\beta = 0.2$&
  \includegraphics[height=0.1\textwidth]{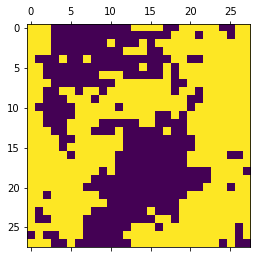} &
   \includegraphics[height=0.1\textwidth]{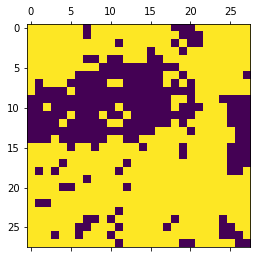} &
  \includegraphics[height=0.1\textwidth]{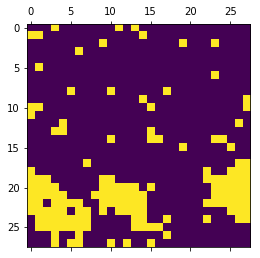} \\
  
  $\beta = 0.3$&
  \includegraphics[height=0.1\textwidth]{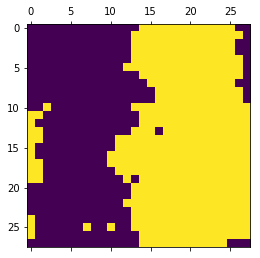} &
  \includegraphics[height=0.1\textwidth]{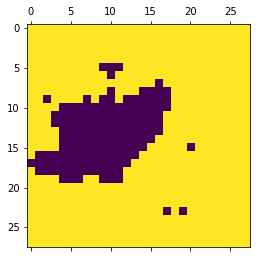} &
  \includegraphics[height=0.1\textwidth]{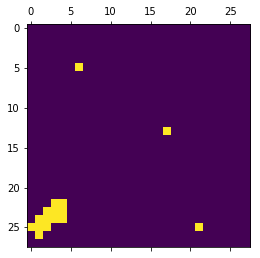} \\
  
  $\beta = 0.4$&
  \includegraphics[height=0.1\textwidth]{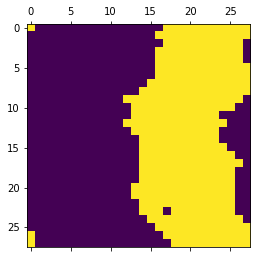} &
  \includegraphics[height=0.1\textwidth]{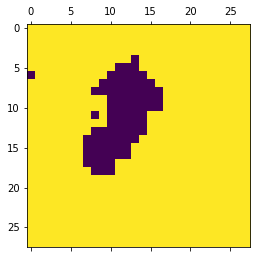} &
  \includegraphics[height=0.1\textwidth]{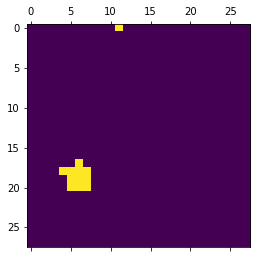} \\
  
  $\beta = 0.5$&
  \includegraphics[height=0.1\textwidth]{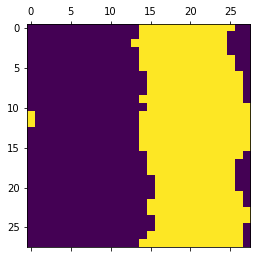} &
  \includegraphics[height=0.1\textwidth]{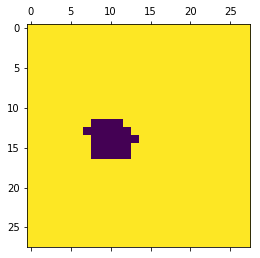} &
  \includegraphics[height=0.1\textwidth]{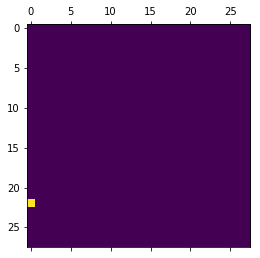} \\
  
  $\beta = 0.6$&
  \includegraphics[height=0.1\textwidth]{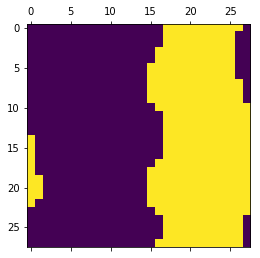} &
  \includegraphics[height=0.1\textwidth]{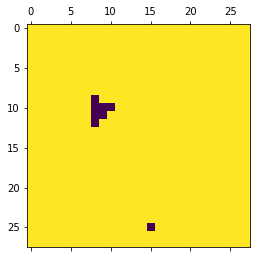} &
  \includegraphics[height=0.1\textwidth]{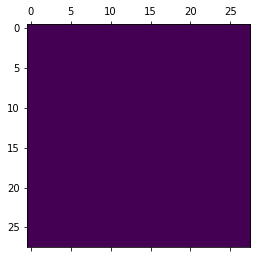} \\
  
  $\beta = 0.7$&
  \includegraphics[height=0.1\textwidth]{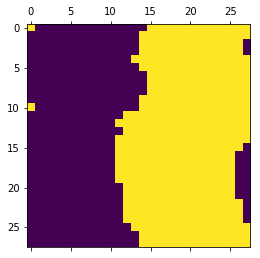} & 
  \includegraphics[height=0.1\textwidth]{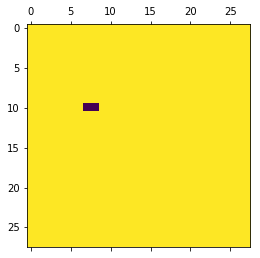} &
  \includegraphics[height=0.1\textwidth]{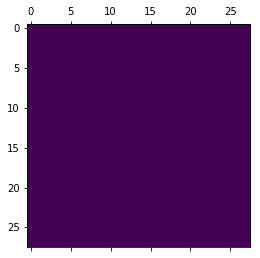} \\
  
  $\beta = 0.8$&
  \includegraphics[height=0.1\textwidth]{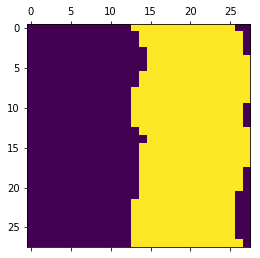} &
  \includegraphics[height=0.1\textwidth]{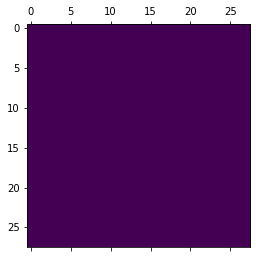} &
  \includegraphics[height=0.1\textwidth]{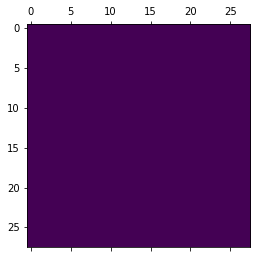} \\
   
  $\beta = 0.9$&
  \includegraphics[height=0.1\textwidth]{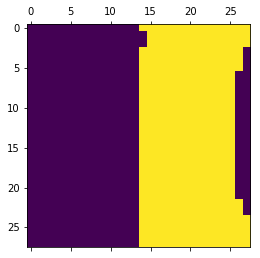} &
  \includegraphics[height=0.1\textwidth]{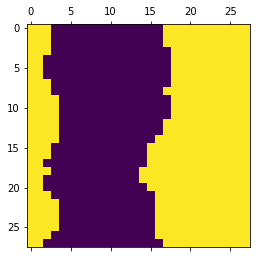} &
  \includegraphics[height=0.1\textwidth]{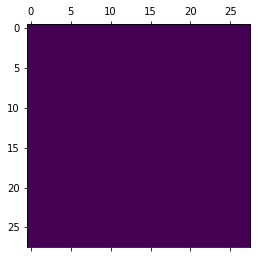} \\
  
  \end{tabular}
  \caption{Examples of images from an Ising model simulation at various $\beta$ parameters.}
  \vspace{-0.15in}
   \label{fig:ising_examples}
\end{figure}

\begin{figure}[ht]
  \centering
  \includegraphics[height=0.3\textwidth]{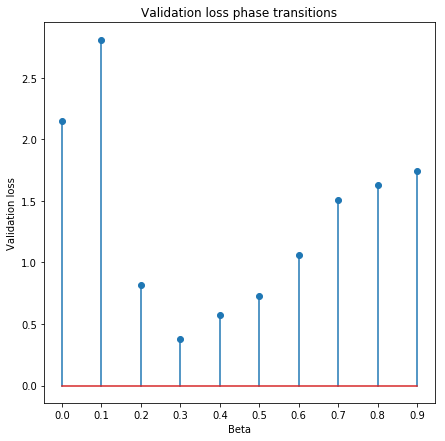}  \\
  \includegraphics[height=0.3\textwidth]{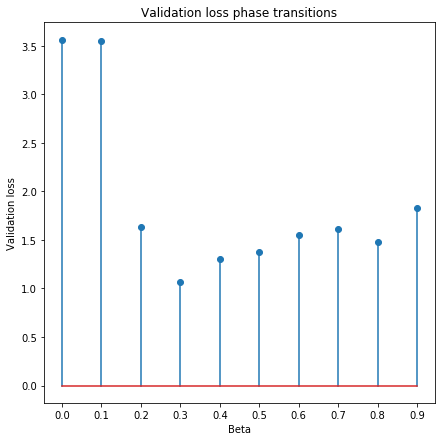} \\ 
  \includegraphics[height=0.3\textwidth]{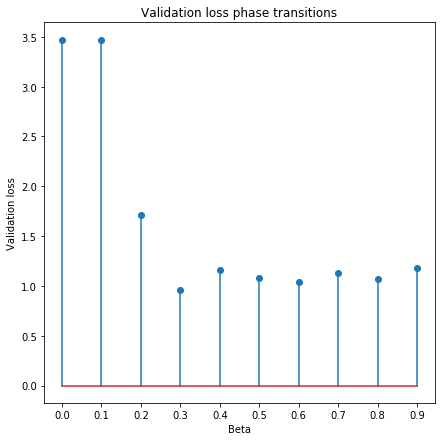} \\
  \includegraphics[height=0.3\textwidth]{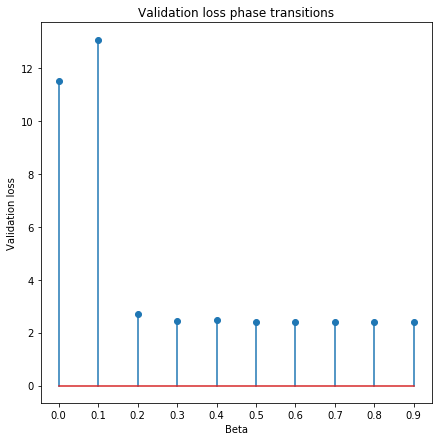}
  \caption{Occam's hill effect for loss when $\beta$ is varied. From top to bottom: MNIST, KMNIST, Fashion MNIST, notMNIST.}
  \vspace{-0.15in}
   \label{fig:ising_loss}
\end{figure}

\begin{figure}[ht]
  \centering
  \includegraphics[height=0.3\textwidth]{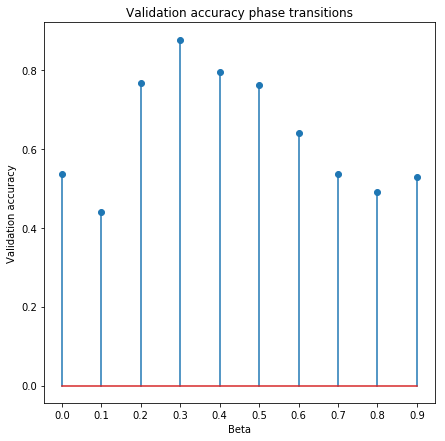}  \\
  \includegraphics[height=0.3\textwidth]{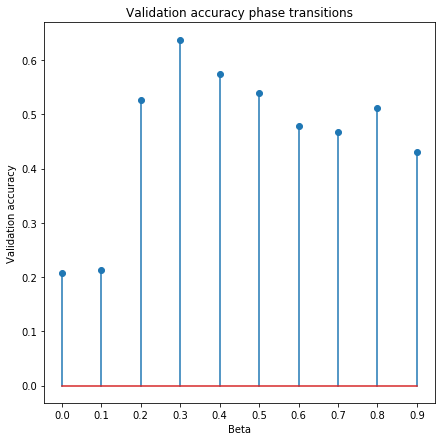} \\ 
  \includegraphics[height=0.3\textwidth]{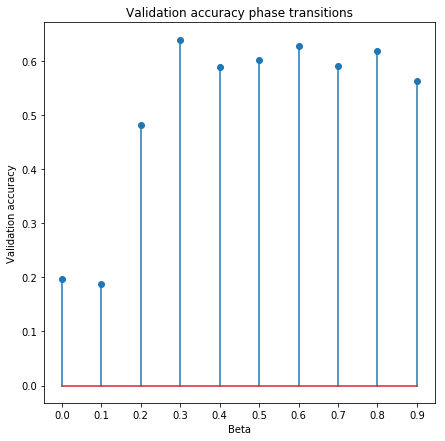} \\
  \includegraphics[height=0.3\textwidth]{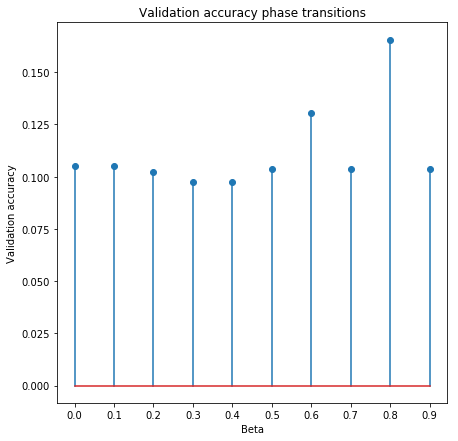}
  \caption{Occam's hill effect for accuracy when $\beta$ is varied. From top to bottom: MNIST, KMNIST, Fashion MNIST, notMNIST.}
  \vspace{-0.15in}
   \label{fig:ising_acc}
\end{figure}

\end{document}